\documentclass[sigconf,letterpaper]{acmart}

\usepackage{balance}
\usepackage{multirow}
\usepackage{enumerate}
\usepackage{pifont}
\usepackage{listings}
\usepackage{algorithm}
\usepackage{algorithmicx}
\usepackage{amsmath}
\usepackage{xcolor}
\usepackage{enumitem}
\usepackage{tcolorbox}
\usepackage{etoolbox}

\definecolor{darkred}{RGB}{139,0,0}

\makeatletter
\AfterEndEnvironment{algorithm}{\let\@algcomment\relax}
\AtEndEnvironment{algorithm}{\kern2pt\hrule\relax\vskip3pt\@algcomment}
\let\@algcomment\relax
\newcommand\algcomment[1]{\def\@algcomment{\footnotesize#1}}
\renewcommand\fs@ruled{\def\@fs@cfont{\bfseries}\let\@fs@capt\floatc@ruled
  \def\@fs@pre{\hrule height.8pt depth0pt \kern2pt}%
  \def\@fs@post{}%
  \def\@fs@mid{\kern2pt\hrule\kern2pt}%
  \let\@fs@iftopcapt\iftrue}
\makeatother

\makeatletter
\def\@fnsymbol#1{\ensuremath{\ifcase#1\or \dagger\or \ddagger\or
   \mathsection\or \mathparagraph\or \|\or **\or \dagger\dagger
   \or \ddagger\ddagger \else\@ctrerr\fi}}
\makeatother

\newcommand{\stitle}[1]{\noindent{\bf #1\/}}

\AtBeginDocument{%
  }

\author{Silin Zhou}
\affiliation{%
  \institution{University of Electronic Science and Technology of China}
  \city{Chengdu}
  \country{China}
}
\email{zhousilinxy@gmail.com}

\author{Chenhao Wang}
\affiliation{%
  \institution{University of Electronic Science and Technology of China}
    \city{Chengdu}
  \country{China}
}
\email{chenhao.wang@std.uestc.edu.cn}

\author{Yuntao Wen}
\affiliation{%
  \institution{University of Electronic Science and Technology of China}
    \city{Chengdu}
  \country{China}
}
\email{yuntaowenx@gmail.com}

\author{Shuo Shang}
\authornote{\authornotemark[1] The corresponding author.}
\affiliation{%
  \institution{University of Electronic Science and Technology of China}
  \city{Chengdu}
  \country{China}
}
\email{jedi.shang@gmail.com}

\author{Lisi Chen}
\affiliation{%
  \institution{University of Electronic Science and Technology of China}
  \city{Chengdu}
  \country{China}
}
\email{lchen012@e.ntu.edu.sg}

\author{Panos Kalnis}
\affiliation{%
  \institution{King Abdullah University of Science and Technology}
  \city{Thuwal}
  \country{Saudi Arabia}
}
\email{panos.kalnis@kaust.edu.sa}

\copyrightyear{2026}
\acmYear{2026}
\setcopyright{cc}
\setcctype{by}
\acmConference[KDD 2026] {Proceedings of the 32nd ACM SIGKDD Conference on Knowledge Discovery and Data Mining V.2}{August 9--13, 2026}{Jeju Island, Republic of Korea.}
\acmBooktitle{Proceedings of the 32nd ACM SIGKDD Conference on Knowledge Discovery and Data Mining V.2 (KDD 2026), August 9--13, 2026, Jeju Island, Republic of Korea}
\acmISBN{979-8-4007-2259-2/2026/08}
\acmDOI{10.1145/3770855.3817701}

\begin{document}

\title{From GPS Points to Travel Patterns: Flexible and Semantic Trajectory Generation with LLMs}


\begin{abstract}
Urban trajectories play a crucial role in modeling urban dynamics and supporting various smart city applications. However, privacy concerns restrict access to large-scale and high-quality trajectory datasets. Trajectory generation provides a promising alternative by synthesizing realistic data to mitigate privacy risks. However, existing methods fail to explicitly capture travel patterns and can only generate fixed-length trajectories under a single condition. To address these limitations, we propose \textbf{HTP}, which \textbf{H}ierarchically generates \textbf{T}ravel patterns first and then generates GPS \textbf{P}oints by using large language models (LLMs), rather than directly generating GPS points. We first design a trajectory-specific residual quantization variational autoencoder (RQ-VAE) that quantizes micro-level GPS trajectories into compact, macro-level travel pattern tokens in a coarse-to-fine manner. These tokens capture rich segment spatial irregularities, such as point density variations caused by traffic conditions. Then, we extend the LLM vocabulary with travel pattern tokens to align trajectory representations with the LLM input, and apply supervised fine-tuning (SFT) to align the LLM with the trajectory generation task, enabling generation of travel pattern sequences under various conditions. Extensive experiments on two real-world datasets show that HTP outperforms the strongest baseline by an average of 29.78\% in terms of generation quality. Our code is available at \url{https://github.com/slzhou-xy/HTP}.
\end{abstract}





\maketitle

\begin{figure}[!t]
    \centering
    \includegraphics[width=0.9\linewidth]{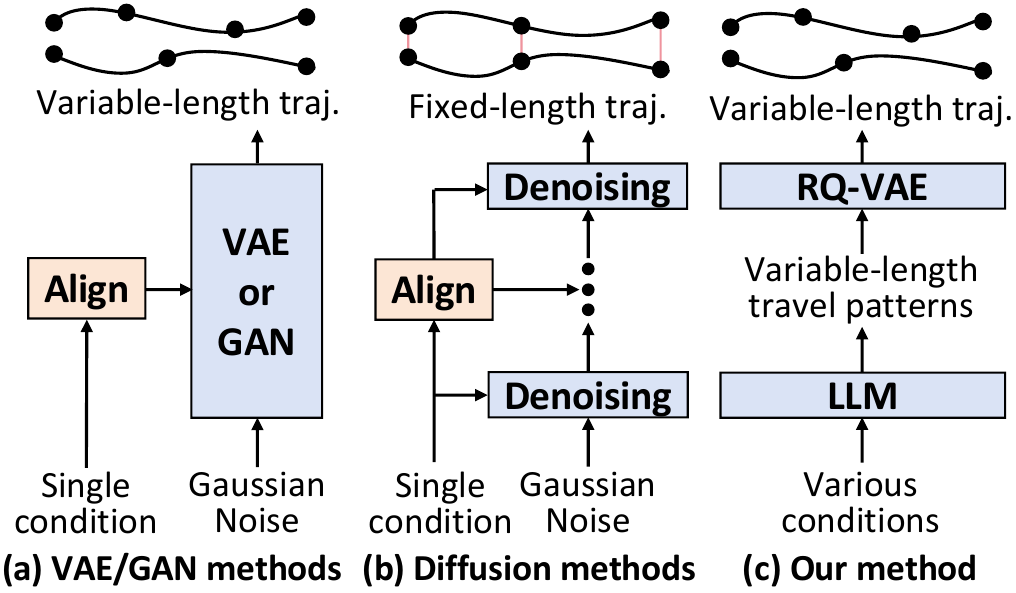}
    \caption{Comparisons of the generation pipelines between existing methods and our proposed method. }
    \label{fig:fig1}
\end{figure}

\section{Introduction}
The widespread utilization of GPS-enabled devices has supported many human mobility tasks, such as urban planning~\cite{urban_planning}, location-based services~\cite{next-poi}, and intelligent transportation~\cite{intelligent_transportation}. 
Mobility trajectory data, which record the spatiotemporal movement, serve as fundamental data for many downstream tasks such as trajectory similarity computation~\cite{traj_sim}, travel time estimation~\cite{travel_time_estimation}, and anomaly detection~\cite{anomaly_detection}. However, access to large-scale real-world trajectory data remains difficult due to privacy concerns~\cite{intro_privacy1} and high data collection costs~\cite{high_data_cost}. Consequently, researchers often lack datasets that capture the richness and complexity of human mobility. To address this issue, trajectory generation~\cite{intro_traj_gen_survey} has been proposed as a promising alternative to provide reliable and privacy-preserving data for research and practical applications.

To generate realistic trajectories, trajectory generation methods should capture two inherent characteristics of real-world mobility trajectories~\cite{intro_traj_survey}: \textit{(1) Complex travel patterns.} Due to variations in travel speed, traffic conditions, and road network structure, some trajectory segments contain dense clusters of GPS points, while others are sparsely sampled~\cite{intro_sim_survey}, resulting in diverse travel patterns. From a macro-level perspective, a trajectory can be decomposed into a sequence of patterns such as acceleration, deceleration, and turning. Even trajectories following the same route may exhibit substantially different compositions of these patterns (due to different start-end times and traffic conditions), which further lead to micro-level variations in point granularity and the number of sampled points. \textit{(2) Various constraints.} Mobility trajectories are influenced by multiple conditions, such as user preferences, travel time, and environmental factors~\cite{foundation_survey}, which jointly determine how a trajectory is generated. For the same origin–destination pair, travelers may select different routes based on traffic or personal preference, creating diverse mobility patterns. Therefore, trajectory generation should satisfy multiple conditional constraints.

Existing trajectory generation methods are based on VAE~\cite{vae,trajvae}, GAN~\cite{gan,trajgan,trajgan_lstm,DP-TrajGAN}, and diffusion~\cite{ddpm,difftraj,controltraj,cardiff} frameworks, as shown in Figure~\ref{fig:fig1}(a)(b). These methods generate trajectories from Gaussian noise. For example, TrajGAN~\cite{trajgan} incorporates LSTMs into GANs~\cite{gan} to model spatial dependencies, while DiffTraj~\cite{difftraj} pioneers the use of diffusion models for progressively denoising and generating trajectories. Despite their effectiveness, these methods still suffer from several limitations. First, they generate trajectories directly at the micro-level (i.e., GPS point) in the implicit embedding space, while overlooking explicit macro-level travel patterns. Therefore, generated trajectories miss important movement characteristics caused by traffic conditions, which manifest as sampling density variations across segments (e.g., congestion leads to dense clusters, while acceleration produces sparse points), causing unrealistic segment-level behaviors. Second, existing methods support only very limited generation conditions and also rely on additional alignment modules to inject conditions into the latent space. Most methods~\cite{trajvae,trajgan,difftraj} only utilize start and end coordinates for the generation condition, ignoring other factors such as travel distance, travel time, and road information. This restricted conditioning limits the diversity and controllability of generation. Moreover, although diffusion-based methods~\cite{ddpm} achieve higher quality generation than VAE-based~\cite{vae} and GAN-based~\cite{gan} methods, they are constrained to fixed-length sequences due to architectural limitations. For example, ControlTraj~\cite{controltraj} fixes the trajectory length of generation to 200 points. Such fixed-length generation deviates from the true length distribution of real-world trajectory data.

To address these issues, we propose a novel two-stage trajectory generation framework, named \textbf{HTP}, as shown in Figure~\ref{fig:fig1}(c). HTP performs hierarchical generation by first producing travel patterns at the macro-level using an LLM~\cite{yang2025qwen3}, and then generating GPS points at the micro-level using an RQ-VAE~\cite{rqvae}. To implement HTP, we tackle two main technical challenges:

\textit{\ding{182} How to transform GPS trajectories into travel patterns?} We design a trajectory-specific RQ-VAE that converts GPS trajectories from free space into shared travel patterns in limited space to capture common behaviors such as congestion and acceleration. The encoder integrates CNNs~\cite{cnn} and Transformers~\cite{transformer} to jointly learn local and global features, and progressively downsamples trajectories to shorten sequence length. This compression effectively learn spatial irregularities from micro-level points to macro-level segment embeddings. We then apply multi-resolution residual quantization to map these embeddings into shared travel patterns. The decoder reconstructs trajectories from the quantized travel patterns together with road contextual information and progressively upsamples the representations. To further capture GPS point density variations, we introduce a relative reconstruction loss.

\textit{\ding{183} How to hierarchically generate variable-length trajectories with various conditions?} 
We use an LLM to first generate macro-level travel patterns and then generate micro-level GPS trajectories.  Instead of alignment modules, we unify various conditions as natural language descriptions, enabling flexible and controllable generation. We extend the LLM vocabulary with travel pattern tokens from the RQ-VAE and perform supervised fine-tuning (SFT) to enable the LLM to generate travel pattern sequences under various conditions. These compressed tokens encode segment-level features, substantially reducing token cost and improving LLM efficiency compared with directly generating GPS points by LLMs. Moreover, the autoregressive structure of LLMs naturally supports variable-length sequence generation. Finally, the generated macro-level travel pattern sequences are fed into the RQ-VAE decoder to generate new variable-length micro-level GPS trajectories.

To evaluate HTP, we conduct experiments on two real-world datasets and compare with 6 SOTA trajectory generation methods. The results show that HTP consistently outperforms the best baseline by an average of 29.78\%. Visualizations confirm HTP can generate high-quality and variable-length trajectories that satisfy the distributions of the real trajectory datasets. An ablation study shows that all designs of HTP contribute to improving performance.

To summarize, we make the following contributions:
\begin{itemize}[leftmargin=*]
    \item We observe that existing trajectory generation methods focus only on micro-level GPS point generation while ignoring macro-level segment travel patterns and support only a single condition, leading to unrealistic and limited diversity generation.
    \item We propose HTP, a novel two-stage framework for hierarchical variable-length trajectory generation. HTP first generates macro-level travel pattern sequences using an LLM, and then generates GPS trajectories from these patterns via an RQ-VAE.
    \item We unify various conditions into natural language descriptions and leverage the world knowledge of LLMs to guide trajectory generation. Combining with travel patterns, HTP enables high-quality and diverse trajectory generation.
\end{itemize}

\begin{figure*}
    \centering
    \includegraphics[width=\linewidth]{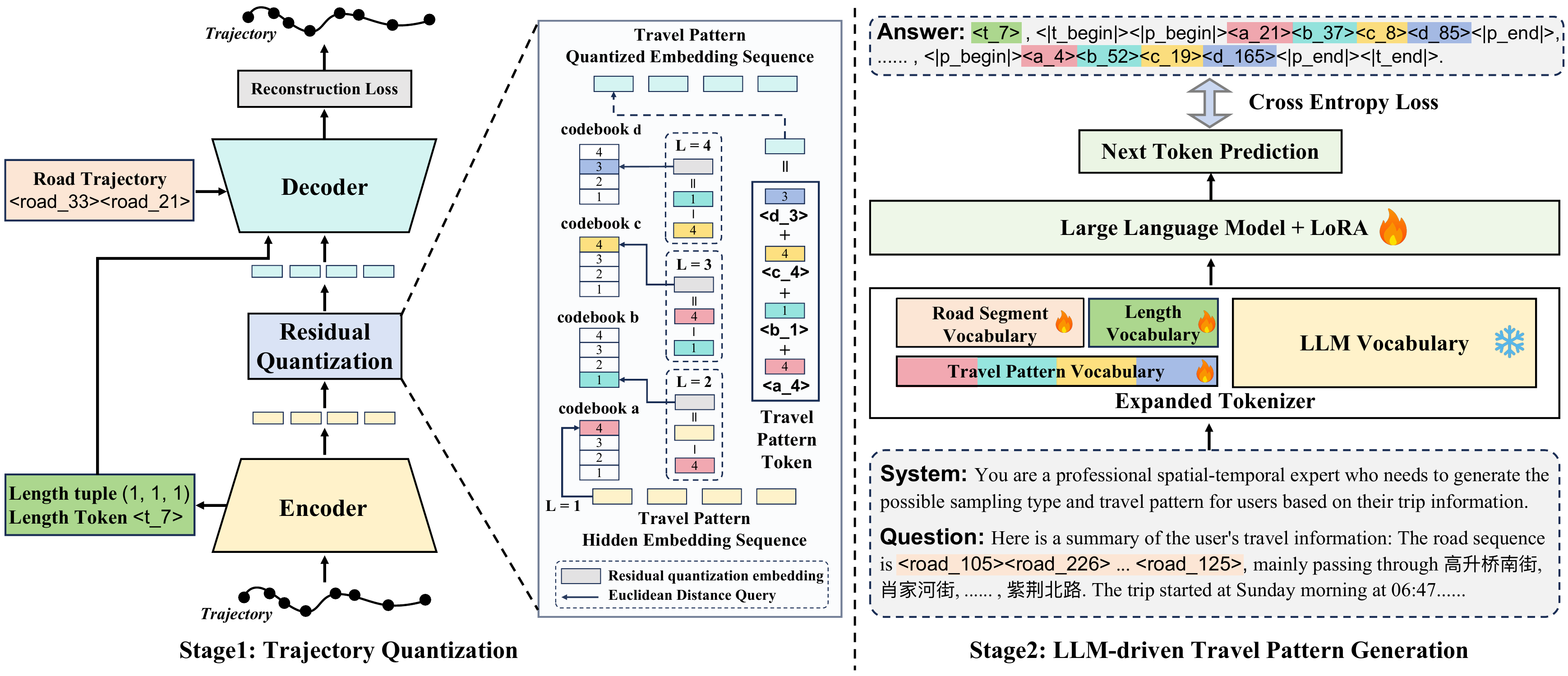}
    \caption{The training overview of HTP. The details of the encoder and decoder are shown in Figure~\ref{fig:encoder_decoder} of the Appendix.}
    \label{fig:framework}
\end{figure*}

\section{Related Work}
\noindent
\textbf{Trajectory Generation.} Trajectory generation aims to synthesize trajectories whose distributions match real-world data~\cite{traj_gen_target,intro_traj_gen_survey}, supporting various downstream applications~\cite{downstream1,downstream2} while mitigating privacy risks~\cite{privacy2}. Early methods rely on rule-based strategies, such as simulations~\cite{simulations} and random perturbations~\cite{perturbations}, but fail to model the complex spatial-temporal dependencies of real trajectories. Recently, learning-based trajectory generation methods~\cite{intro_traj_gen_survey} are proposed to model complex distributions, including VAE-based~\cite{vae}, GAN-based~\cite{gan}, and diffusion-based~\cite{ddpm} frameworks. For example, TrajVAE~\cite{trajvae} introduces LSTM~\cite{lstm} into VAE to learn sequential information. To enhance realism,  GAN-based methods~\cite{DP-TrajGAN,ST-TrajGAN,trajgan,trajgan_lstm} further introduce discriminators. For example, TrajGAN-LSTM~\cite{trajgan_lstm} integrates LSTMs with GANs and a trajectory similarity loss to capture spatiotemporal patterns. However, GANs are notoriously difficult to train and prone to collapse~\cite{gan_collapse}. Recent SOTA methods~\cite{difftraj,controltraj,cardiff} adopt diffusion models, which generate trajectories by progressively denoising Gaussian noise under given conditions. DiffTraj~\cite{difftraj} pioneers diffusion-based trajectory generation using a UNet~\cite{unet} for spatiotemporal denoising, while ControlTraj~\cite{controltraj} incorporates road topology to improve realism. Cardiff~\cite{cardiff} further proposes a two-stage diffusion pipeline that first generates road embeddings and then GPS trajectories. However, diffusion models require fixed-length inputs, limiting variable-length trajectory generation in real-world settings. Unlike existing methods, our method generates travel patterns under various conditions, enabling diverse variable-length trajectory generation.

\stitle{Large Language Models.} 
Since the publication of ChatGPT~\cite{chatgpt}, large language models (LLMs) have rapidly reshaped how humans interact with information~\cite{LLM_survey}. Open-source models such as the LLaMA series~\cite{touvron2023llama,touvron2023llama2,llama3} have enabled strong foundational capabilities and accelerated community-driven research and downstream adaptation. Building on this progress, the Qwen series~\cite{bai2023qwen,yang2024qwen2,yang2025qwen3} improves multilingual support and long-context modeling, making LLMs more practical and scalable for real-world applications. More recently, DeepSeek-R1~\cite{ds-r1} represents a new generation of large reasoning models (LRMs), significantly reducing training cost through the proposed GRPO~\cite{grpo} reinforcement learning algorithm while maintaining competitive performance. In this work, we unify various conditions into textual descriptions and leverage LLM world knowledge to generate diverse travel patterns.

\section{Problem Definition} \label{sec:problem_formulation}

\noindent
\textbf{Definition1 (GPS Trajectory)}. A GPS trajectory is defined as a variable-length sequence of continuously sampled GPS points, denoted as $\tau = \langle p_i \rangle_{i=1}^{n}$ with length $n$, collected under a fixed sampling frequency. Each point $p_i \in \tau$ is represented as a tuple $(x_i, y_i)$, where $x_i$ and $y_i$ denote the longitude and latitude.

\stitle{Definition2 (Road Network)}. A road network is represented as a directed graph $G = (V, A)$, where $V$ is the set of road segments and $A \in \mathbb{R}^{|V| \times |V|}$ is the adjacency matrix indicating their connectivity. Specifically, $A[i, j] = 1$ if road segments $v_i$ and $v_j$ are directly connected, and $A[i, j] = 0$ otherwise. Each road segment $v_i$ is associated with a GPS point sequence $S_i = \langle p_i \rangle_{i=1}^{|S_i|}$ representing its geometry linestring.

\stitle{Definition3 (Road Trajectory)}. A road trajectory $\tau_{r}=\langle v_i \rangle_{i=1}^{n_r}$ is a sequence of road segments with length $n_r$ that is generated from a GPS trajectory $\tau$ using map-matching algorithms, where $v_i$ denotes the ID of a road segment in the road network $G$.

\stitle{Problem Definition (Mobility Trajectory Generation)}. Given a real-world GPS trajectory dataset $\mathcal{D} = \{ \tau_i \}_{i=1}^{|\mathcal{D}|}$, the mobility trajectory generation task aims to synthesize a new high-quality and diverse mobility GPS trajectory dataset $\widetilde{\mathcal{D}} = \{ \widetilde{\tau}_i\}_{i=1}^{|\widetilde{\mathcal{D}}|}$ under the provided conditions. The generated trajectory dataset $\widetilde{\mathcal{D}}$ has a similar distribution and travel pattern to the original dataset $\mathcal{D}$.
\section{The Proposed Method HTP}
Figure~\ref{fig:framework} presents the training architecture of HTP, which consists of two stages: \textit{Stage 1: Trajectory Quantization.} A trajectory-specific RQ-VAE compresses GPS trajectories into shared travel pattern tokens from micro-level to macro-level, transforming point sequences into segment tokens.
\textit{Stage 2: LLM-driven Travel Pattern Generation.} An LLM with expanded vocabulary is fine-tuned to generate quantized travel pattern tokens under various conditions.

\subsection{Trajectory Quantization}
Raw GPS points describe trajectories only at the micro-level and carry limited spatial information, making it difficult to capture the complex spatial irregularities of real-world movement. In contrast, important macro-level information, such as segment-level travel patterns, can reflect the local movement from point density variations. To achieve this, we design a trajectory-specific RQ-VAE in the first stage to explicitly quantize GPS trajectories from free space into shared, multi-resolution travel pattern tokens in a limited space for modeling common macro-level movement.

\stitle{Travel Pattern Encoder.} The encoder learns trajectory segment embeddings of the macro-level from raw GPS trajectories. Formally, we learn trajectory segment embeddings for a normalized GPS trajectory $\tau=\left< p_i\right>_{i=1}^{n}$ as follows:
\begin{equation}
\begin{aligned}
    \mathbf{H} &= \mathtt{Linear}(\tau) \in \mathbb{R}^{n \times d}, \\
    \mathbf{H}_e &= \mathtt{Encoder}(\mathbf{H}) \in \mathbb{R}^{\frac{n}{8} \times d_e},
    \label{eq:encoder}
\end{aligned}
\end{equation}
where the encoder progressively 2$\times$downsamples the trajectory point embeddings $\mathbf{H}$ to compress the sequence while doubling the embedding dimensionality, enabling the extraction of rich macro-level segment information. The encoder alternates between CNN and Transformer layers to jointly capture local and global travel movement of GPS trajectories. After three downsampling steps, the sequence length is reduced to $\frac{n}{8}$. This compression substantially shortens the sequence while increasing feature dimensionality, thereby preserving essential information for modeling travel patterns at the segment-level. Here, sequence padding is used to handle variable-length inputs for both CNNs and Transformers.

Since GPS trajectories have variable lengths, sequence lengths will change irregularly during progressive downsampling. When a trajectory length is a power of 2, e.g., $n=32$, three $2\times$ downsamplings $32\rightarrow 16 \rightarrow 8 \rightarrow 4$ and $2\times$upsamplings of the decoder can perfectly restore the original sequence length. However, for non-power-of-2 lengths, e.g., $n=26$, rounding introduces odd–even fluctuations $26 (\text{even}) \rightarrow 13 (\text{odd}) \rightarrow 7 (\text{odd}) \rightarrow 4 (\text{even})$. When 2$\times$upsampling, the reconstructed sequence cannot match the original length, e.g., $4 \rightarrow \underline{8} \rightarrow \underline{16} \rightarrow \underline{32}$.

To address this issue, we explicitly record the odd–even transformation of sequence lengths during each downsampling step and provide it to the decoder for correction. Specifically, we use 1 to indicate an odd length that requires adjustment and 0 to indicate an even length that does not. For example, $26 (\underline{0}) \rightarrow 13 (\underline{1}) \rightarrow 7 (\underline{1}) \rightarrow 4 $ is recorded as $(0,1,1)$. During upsampling, the decoder reverses this process using $4 \rightarrow 7(4\times2-\underline{1}) \rightarrow 13 (7\times2-\underline{1}) \rightarrow 26 (13\times2-\underline{0})$, ensuring accurate sequence length recovery.

\stitle{Multi-resolution Quantization.} Although trajectory segment embeddings capture macro-level information, they are learned independently for each trajectory and therefore cannot represent common travel patterns shared across different trajectories, such as deceleration or acceleration segments caused by congestion. To address this issue, we introduce quantization to convert these independent segment embeddings into shared travel pattern tokens. Specifically, we first apply a linear projection to reduce the embedding dimensionality and distill core travel pattern information:
\begin{equation}
    \mathbf{H}_q = \mathtt{Linear}(\mathbf{H}_e) \in \mathbb{R}^{\frac{n}{8} \times d_q},
\end{equation}
where $d_q < d_e$. Then, we convert $\mathbf{h}_{q, i} \in \mathbf{H}_q$ into a shared travel pattern token by vector quantization (VQ).
The token is obtained by the Euclidean distance query between $\mathbf{h}_{q,i}$ and all embeddings in the shared quantization codebook
$\mathbf{E} \in \mathbb{R}^{C \times d_q}$ as follows:
\begin{equation}
\begin{aligned}
    k_{i} &=  \underset{k \in \{1,\dots,C\}}{\mathtt{argmin}} \|\mathbf{h}_{q,i} - \mathbf{e}_k\|_2, \\
    K &= (k_1, k_2, \dots, k_{\frac{n}{8}}), \\
    \mathbf{E}_q &= [\mathbf{e}_{k_1}, \mathbf{e}_{k_2}, \dots, \mathbf{e}_{k_{\frac{n}{8}}}]\in\mathbb{R}^{\frac{n}{8} \times d_q},
\end{aligned}
\label{eq:quant}
\end{equation}
where $\mathbf{e}_k \in \mathbf{E}$, $K$ is the index tuple recording the token IDs of the nearest embeddings in $\mathbf{E}$, e.g., travel pattern ID sequence $K = (16, 5, \dots, 24)$ for the GPS trajectory $\tau$, and $\mathbf{E}_q$ denotes the quantized embeddings obtained by indexing $\mathbf{E}$ with $K$.

Since single-layer quantization provides limited expressiveness, we adopt the $L$-layer residual quantization (RQ). Specifically, we employ $L$ shared quantization codebooks with sizes $(C_1, C_2, \dots, C_L)$. For each layer $l \in \{1, \dots, L\}$, we first apply Equation~\ref{eq:quant} to obtain the quantized embeddings $\mathbf{E}_q^{(l)}$ and then compute residuals as:
\begin{equation}
    \mathbf{H}^{(l + 1)}_q = \mathbf{H}^{(l)}_q - \mathbf{E}_q^{(l)},
\end{equation}
where $\mathbf{H}_q^{(1)} = \mathbf{H}_q$. After $L$ RQs, we obtain the multi-resolution travel pattern token $\mathcal{K} =[K^{(1)}, K^{(2)}, \dots, K^{(L)}] \in \mathbb{R}^{L \times \frac{n}{8}}$ and the quantized embeddings $[\mathbf{E}_q^{(1)}, \mathbf{E}_q^{(2)}, \dots, \mathbf{E}_q^{(L)}] \in \mathbb{R}^{L \times \frac{n}{8} \times d_q}$ for the trajectory $\tau$. The overall quantized embedding is computed as: 
\begin{equation}
    \mathbf{H}_q^\prime = \mathtt{Linear}(\sum_{l=1}^{L}\mathbf{E}_q^{(l)}) \in \mathbb{R}^{\frac{n}{8} \times d_e},
\label{eq:quant_sum}
\end{equation}
Here, $\mathbf{H}_q$ can be approximated by $\sum_{l=1}^{L}\mathbf{E}_q^{(l)}$, i.e, $\mathbf{H}_q\approx\sum_{l=1}^{L}\mathbf{E}_q^{(l)}$. 

By employing RQ, the expressiveness of the token space is expanded to $\prod_{l=1}^{L} C_l$, indicating that HTP can represent up to $\prod_{l=1}^{L} C_l$ distinct types of travel patterns. To learn the coarse-to-fine characteristics of spatial information, we set $C_1 \textless C_2\textless \cdots \textless C_L$ in RQ, where the shallower layers capture coarse-grained travel semantics and the deeper layers refine them into fine-grained patterns. 

\stitle{Travel Pattern Decoder.} The decoder reconstructs micro-level trajectory from the quantized embeddings $\mathbf{H}_q^\prime$ of the macro-level by three $2\times$upsampling steps, each composed of CNN and Transformer layers. Formally, this process is defined as:
\begin{equation}
    \mathbf{H}^\prime = \mathtt{Decoder}(\mathbf{H}_q^\prime) \in \mathbb{R}^{n \times d},
\end{equation}
where $\mathbf{H}^\prime$ is the reconstructed trajectory embedding.

However, quantization compresses GPS trajectories into discrete travel pattern sequences, inevitably losing spatial information. To solve this, we incorporate the corresponding road trajectory $\tau_r$ as external context and also learn point density along road segments. Specifically, we use Node2vec~\cite{node2vec} to learn road segment embeddings that capture relative spatial relationships on road network $G$, and transform $\tau_r$ into a road embedding sequence $\mathbf{H}_g \in \mathbb{R}^{n_r \times d_r}$. We further leverage the normalized GPS linestring geometry $S_i$ of each road segment $v_i$ to learn absolute spatial embedding $\mathbf{H}_s$:
\begin{equation}\label{eq:road_pooling}
\begin{aligned}
    \mathbf{h}_{s,i} &= \mathtt{MeanPooling}\big(\mathtt{Linear}(S_i)\big) \in \mathbb{R}^{d_r}, \\
    \mathbf{H}_s &=[\mathbf{h}_{s,1}, \mathbf{h}_{s,2}, ..., \mathbf{h}_{s,n_r}] \in \mathbb{R}^{n_r \times d_r},
\end{aligned}
\end{equation}
The final road sequence embedding is $\mathbf{H}_r = \mathbf{H}_g + \mathbf{H}_s$. We further refine road sequence embedding by a Transformer encoder:
\begin{equation}\label{eq:road_trm}
\mathbf{H}_r^\prime = \mathtt{Transformer}(\mathbf{H}_r + \mathbf{P}) \in \mathbb{R}^{n_r \times d_r},
\end{equation}
where $\mathbf{P}$ is the positional encoding~\cite{transformer}. Finally, $\mathbf{H}_r^\prime$ is fed into all Transformer layers of the decoder by adding cross-attention. 

\stitle{Loss Functions.} In the RQ-VAE, the overall loss consists of two parts: quantization loss and trajectory reconstruction loss. The quantization loss is defined as:
\begin{equation}
\mathcal{L}_{q} = \frac{1}{L} \sum_{l = 1}^{L}
\Big(
\parallel\mathtt{sg}(\mathbf{H}_q^{(l)}) - \mathbf{E}_q^{(l)}\parallel^2_2 + \beta \parallel \mathbf{H}_q^{(l)}- \mathtt{sg}(\mathbf{E}_q^{(l)})\parallel^2_2
\Big),
\end{equation}
where $\mathtt{sg}(\cdot)$ denotes the stop-gradient operation because of $\mathtt{argmin}$ operation in Equation~\ref{eq:quant}, and $\beta$ is a balancing coefficient controlling the commitment cost between the encoder and the codebook.

\begin{figure}
    \centering
    \includegraphics[width=0.9\linewidth]{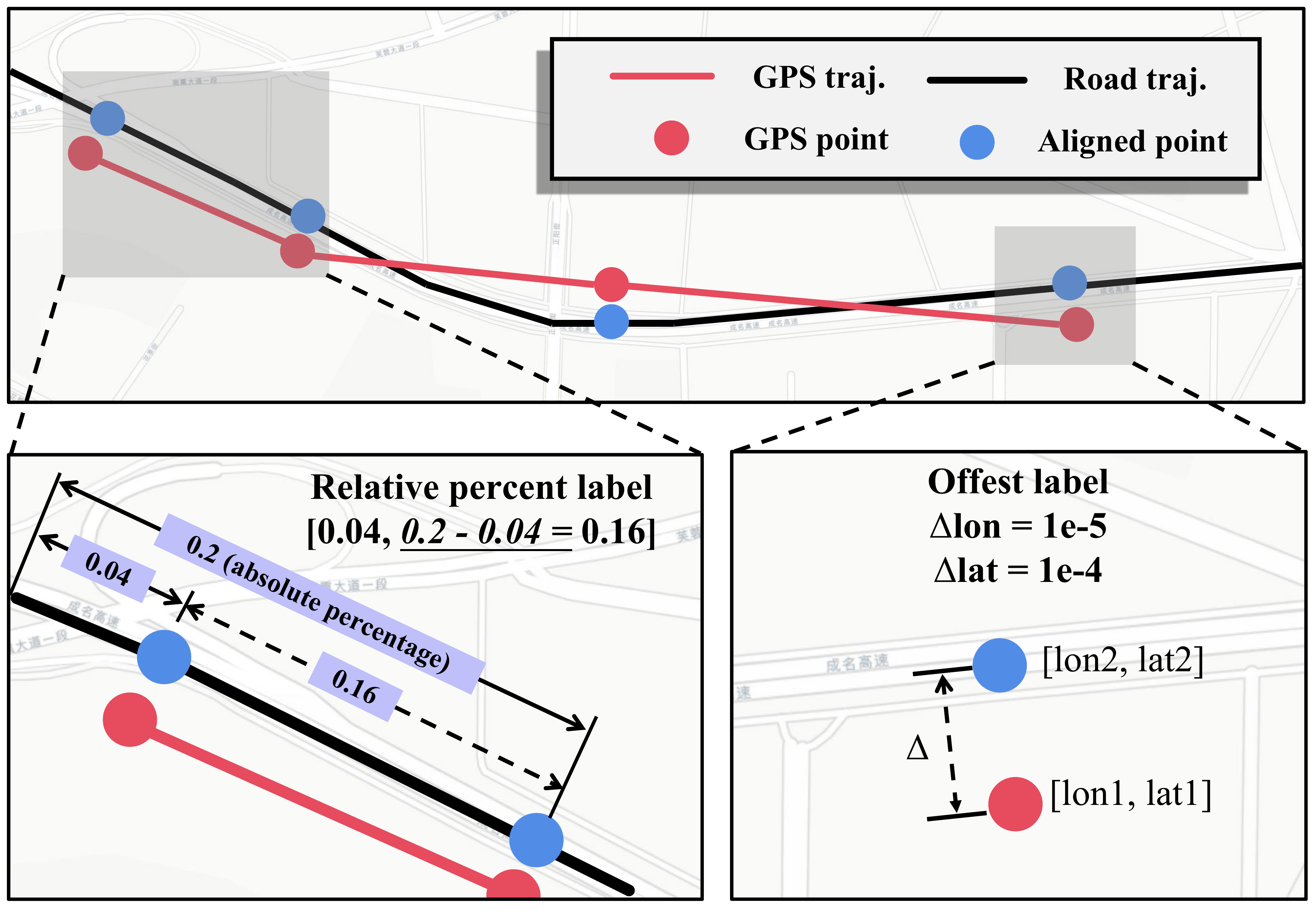}
    \caption{Illustration of the process for obtaining training labels for our relative reconstruction loss.}
    \label{fig:loss}
\end{figure}

Previous studies~\cite{difftraj,controltraj,cardiff} typically reconstruct trajectories on GPS points in free space. However, due to the extremely high precision of GPS coordinates, even small reconstruction errors can cause large spatial deviations, making optimization difficult. To address this issue, we propose \textit{relative reconstruction loss} that guides trajectory reconstruction on road segments. Figure~\ref{fig:loss} shows the process of obtaining the training labels. Specifically, each red point of a real trajectory can be aligned to a blue point of the corresponding road segment of a road trajectory (black line). Therefore, we can reconstruct this real trajectory by locating its point on a road segment with coordinates offset caused by GPS-enabled devices.  

\textit{(1) Relative Percent Loss.}
This loss computes the relative position of each GPS point along the road trajectory. As illustrated in the bottom-left of Figure~\ref{fig:loss}, we first compute the absolute percentage, ranging from $[0,1]$, representing the position of each blue point relative to the start of the road trajectory. We then convert these values into relative percentages by measuring each blue point’s increment with respect to the previous one. These relative percentages capture the point density variations on road segments. Given the ground-truth percentage values $Y_\mathrm{percent}$, the loss is defined as:
\begin{equation}
\mathcal{L}_\mathrm{percent} = \mathtt{MSE}\big(Y_\mathrm{percent}, \mathtt{MLP}(\mathbf{H}^\prime)\big),
\end{equation} 
where $\mathtt{MLP}$ predicts the position percent within a road segment.

\textit{(2) Offset Loss.}
In reality, GPS points may not lie on road segment and exhibit small deviations due to GPS-enabled device noise and environmental factors. To model this deviation, we introduce an offset loss. As shown in the bottom-right of Figure~\ref{fig:loss}, the offset label is obtained by computing the displacement between each real GPS point and its aligned point on the road segment. Given the ground-truth offset values $Y_\mathrm{offset}$, the loss is defined as:
\begin{equation}
\mathcal{L}_\mathrm{offset} = \mathtt{MSE}\big(Y_\mathrm{offset}, \mathtt{MLP}(\mathbf{H}^\prime)\big),
\end{equation}
where $\mathtt{MLP}$ predicts the offset. Here, the ground-truth offset labels are normalized, as their raw magnitudes are extremely small, e.g., $\Delta_\mathrm{lon} \approx 10^{-5}$, $\Delta_\mathrm{lat} \approx 10^{-4}$.

Unlike reconstruction over the entire map space, our relative loss reconstructs trajectories within a constrained local space, and their targets lie in narrow and fixed ranges, making the optimization both stable and spatially bounded. 

The reconstruction loss is $\mathcal{L}_{r} = \mathcal{L}_\mathrm{percent} + \mathcal{L}_\mathrm{offset}$. Therefore, the final training loss for the trajectory-specific RQ-VAE is:
\begin{equation}
\mathcal{L} = \mathcal{L}_{q} + \mathcal{L}_{r}.
\end{equation}

\begin{figure}
    \centering
    \includegraphics[width=\linewidth]{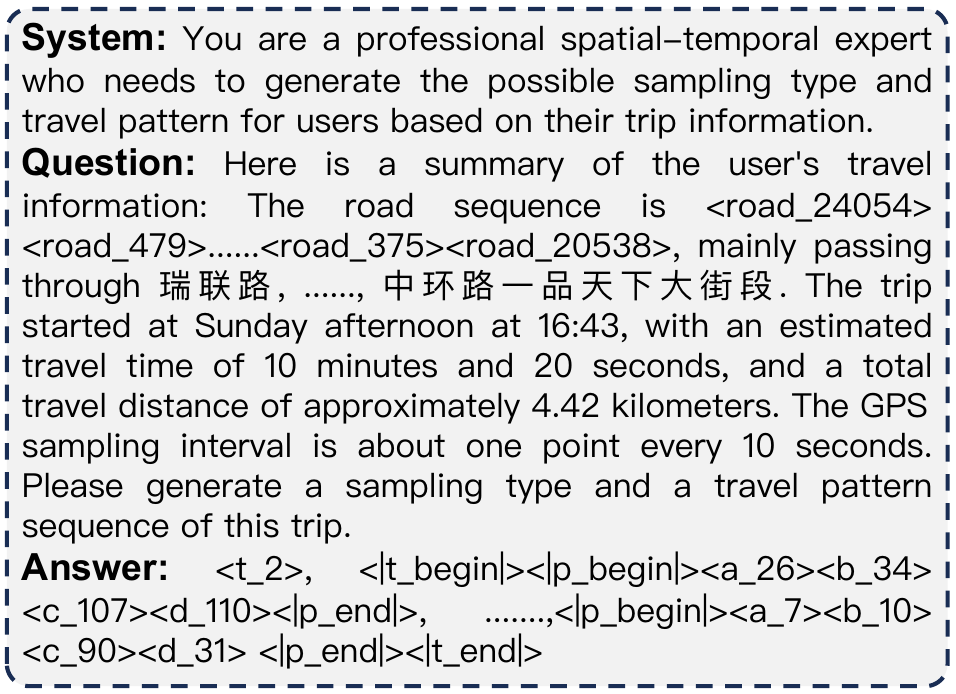}
    \caption{The question-answer format on Chengdu dataset.}
    \label{fig:prompt}
\end{figure}

\subsection{LLM-driven Trajectory Generation}
Existing trajectory generation methods can only leverage limited conditions with additional alignment modules, such as start and end points~\cite{intro_traj_gen_survey}. However, real-world mobility trajectory is influenced by various factors~\cite{intro_traj_survey}. In addition, diffusion-based methods~\cite{cardiff} are limited to fixed-length inputs, while autoregressive methods~\cite{trajgan_lstm} support variable lengths but directly generate GPS points, leading to cumulative spatial drift.

To address these issues, we introduce an LLM-driven generation in the second stage, as illustrated in Figure~\ref{fig:framework}. Various conditions, such as road segment IDs, travel time, and so on, are converted into textual descriptions, providing a unified, flexible, and interpretable conditioning interface. We then perform SFT on an LLM with question–answer pairs. The complete format of a question–answer pair is provided in the Figure~\ref{fig:prompt}. Instead of directly generating GPS points on the micro-level, HTP generates travel pattern tokens and length tokens derived from the RQ-VAE. By leveraging the LLM’s rich world knowledge, HTP can produce high-quality, variable-length trajectories under various conditions.

To align road segments, travel patterns, and length records to the LLM, we expand the LLM’s vocabulary with specialized tokens. Each road segment is represented as a road token $\mathtt{\langle road\_ID \rangle}$, yielding a total of $V$ road tokens. Travel-pattern tokens, defined as $\mathtt{\langle a\_ID \rangle}$, $\mathtt{\langle b\_ID \rangle}$, etc., correspond to the quantization layers in the RQ-VAE with total $\sum_{l=1}^{L} C_l$ tokens. Length tokens $\mathtt{\langle t\_ID \rangle}$ encode the length records produced by the encoder. Because three downsampling layers introduce odd–even variations, we obtain $2^3 = 8$ possible length tokens. For instance, $\mathtt{\langle t\_7 \rangle}$ corresponds to the length record $(1,1,1)$. The full transformation rules for these records are provided in Figure~\ref{fig:length_token} of the Appendix. Additionally, we introduce four special tokens, i.e., $\mathtt{\langle\vert t\_begin \vert\rangle}$, $\mathtt{\langle\vert t\_end \vert\rangle}$, $\mathtt{\langle\vert p\_begin \vert\rangle}$, and $\mathtt{\langle\vert p\_end \vert\rangle}$, to mark the boundaries of travel pattern sequences. In total, the expanded vocabulary includes $V + \sum_{l=1}^{L} C_l + 8 + 4$ tokens. We apply LoRA~\cite{lora} to fine-tune the LLM and incorporate these additional tokens effectively.

When generating new trajectories, the LLM first generates travel pattern tokens and a length token by condition descriptions on the macro-level. The generated tokens are mapped to quantized embeddings using codebooks and aggregated according to Equation~\ref{eq:quant_sum}. The RQ-VAE decoder then generates relative positions and spatial offsets conditioned on the road segments, which are finally combined to generate the final GPS trajectory on the micro-level.

\begin{table*}[!t]
 \LARGE
	\centering
	\caption{Results of HTP and the baselines on Chengdu and Porto datasets. The best results are highlighted in bold, the second-best results are \underline{underlined}, and the bottom row is the relative improvement of HTP over the best baseline. 
    }
	\resizebox{\linewidth}{!}{
		\begin{tabular}{l|cccccccc|cccccccc}
			\toprule
          & \multicolumn{8}{c}{\textbf{Chengdu}} & \multicolumn{8}{c}{\textbf{Porto}}
          \\\cmidrule(lr){2-9}\cmidrule(lr){10-17}
          & \multicolumn{3}{c}{\textbf{Point-level}} & \multicolumn{2}{c}{\textbf{Grid-level}}
          & \multicolumn{3}{c}{\textbf{Road-level}}
          & \multicolumn{3}{c}{\textbf{Point-level}} & \multicolumn{2}{c}{\textbf{Grid-level}}
          & \multicolumn{3}{c}{\textbf{Road-level}}
       \\\cmidrule(lr){2-4}\cmidrule(lr){5-6}\cmidrule(lr){7-9}\cmidrule(lr){10-12}
       \cmidrule(lr){13-14}\cmidrule(lr){15-17}
			& T-Dist$\downarrow$   
            & S-Dist$\downarrow$  
			& Radius$\downarrow$ 
            & G-Den$\uparrow$  
            & G-Pat$\uparrow$  
            & R-Den$\uparrow$  
            & R-Pat$\uparrow$  
            & PR-Dist$\downarrow$
			& T-Dist$\downarrow$   
            & S-Dist$\downarrow$  
			& Radius$\downarrow$  
            & G-Den$\uparrow$  
            & G-Pat$\uparrow$  
            & R-Den$\uparrow$  
            & R-Pat$\uparrow$  
            & PR-Dist$\downarrow$
			\\
			\midrule
			TrajGAN 
			& 0.02008  & 0.00914  & 0.00248     
			& 0.57888  & 0.14286  
			& 0.56407 & 0.09059 & 0.42391
            & 0.07001  & 0.02796 & 0.06960
            & 0.70959 & 0.24719
            & 0.65745 & 0.23881 & 0.36468
			\\   
            TrajVAE
            & 0.01168  & 0.04222  & 0.00228     
			& 0.63091  & 0.14286  
			& 0.62535 & 0.04778 & 0.43777
            & 0.03884  & 0.06019  &  0.00327    
			& 0.73015  & 0.25564  
			& 0.71300 & 0.38095 & 0.40190
			\\   
            DiffWave
            & 0.00055  & \underline{0.00237}  & 0.00152     
			&  0.92800 & 0.54237  
			& 0.94388 & 0.44898 & 0.25695
            & \underline{0.00034}  & 0.00622  & 0.00487     
			& 0.97020  & 0.62009  
			& 0.97422 & 0.77778 & 0.09579 
			\\ 
            DiffTraj
            & 0.00052  & 0.00387  & 0.00028     
			& 0.95136  & 0.67857  
			& 0.94949 & 0.52101 & 0.24121
            & 0.00093  & 0.00587  & 0.00100     
			& 0.96838  &  0.58621 
			& 0.97654 & 0.73973 & 0.13961
			\\ 
            ControlTraj
            & \underline{0.00032}  & 0.00920  & \underline{0.00021}   
			& \underline{0.97274}  & \underline{0.75229}
			& \underline{0.96457} & \underline{0.55319} & \underline{0.15164}
            & 0.00066  & \underline{0.00529}  & 0.00086
			& 0.97733  & 0.62009  
			& 0.98003 & 0.83328 & 0.06777 
			\\ 
            Cardiff
            & 0.00070  & 0.02399  & 0.00025     
			& 0.94349  &  0.54237 
			& 0.94140 & 0.39044 & 0.21681
            & 0.00279  & 0.01365  & \underline{0.00019}     
			& \underline{0.98196}  & \underline{0.69058}
			& \underline{0.98641} & \underline{0.87081} & \underline{0.05993}
			\\ 
			\midrule
            \textbf{HTP(ours)}
            & \textbf{0.00027}  & \textbf{0.00069}  & \textbf{0.00019}     
			& \textbf{0.99567}  & \textbf{0.84360}  
			& \textbf{0.99731} & \textbf{0.94118} & \textbf{0.02955}
            & \textbf{0.00020}  & \textbf{0.00232}  & \textbf{0.00016}     
			& \textbf{0.99519}  & \textbf{0.83019}  
			& \textbf{0.99418} & \textbf{0.92683} & \textbf{0.01799} 
			\\ 
            \midrule
			\textbf{Improve.}
			& \textbf{15.63\%}  & \textbf{70.89\%}  & \textbf{9.52\%}     
			& \textbf{2.36\%}  & \textbf{12.14\%} 
			& \textbf{3.39\%}  & \textbf{70.14\%}  & \textbf{80.51\%} 
			& \textbf{41.18\%}  & \textbf{56.14\%}  & \textbf{15.79\%}     
			& \textbf{1.35\%}  & \textbf{20.22\%} 
			& \textbf{0.79\%} & \textbf{6.43\%}  & \textbf{69.98\%}
			\\
			\bottomrule
		\end{tabular}
	}
	\label{tab:main_results}
\end{table*}

\stitle{Limitations and Complexity Analysis.} Since HTP incorporates LLMs, it introduces additional computational overhead compared with VAE-based, GAN-based, and diffusion-based methods. To mitigate this cost, LLM acceleration frameworks, such as vLLM~\cite{vllm} and SGLang~\cite{SGLang}, can be employed. Moreover, the compression of trajectory-specific RQ-VAE substantially reduces the generation cost of the LLM. Specifically, suppose a GPS trajectory contains $n$ points and each coordinate is represented with five decimal digits. After LLM tokenization, directly representing a GPS trajectory results in about $15n$–$20n$ tokens. In contrast, our trajectory-specific RQ-VAE compresses a GPS trajectory into a travel pattern sequence of length approximately $\frac{n}{8}$. Consequently, compared with directly generating GPS point using an LLM, HTP reduces the number of generated tokens by nearly two orders of magnitude, significantly improving generation efficiency. Consequently, despite involving LLMs, HTP is not substantially slower than existing methods. Detailed efficiency analysis is shown in Experiment~\ref{app:inference_time}.

 \section{Experimental Evaluation} \label{section:exp}
We experiment extensively to answer the following questions:
\begin{itemize}[leftmargin=*]
    \item \textbf{RQ1}: How well does HTP generate trajectory quality and realism compared to SOTA methods?
    \item \textbf{RQ2}: How do the key design components of HTP contribute to the overall generation quality?
    \item \textbf{RQ3}: How efficient is HTP on trajectory generation?
    \item \textbf{RQ4}: How do different hyperparameter settings influence the quality of generated trajectories?
    \item \textbf{RQ5}: What representations does the RQ-VAE learn, and how does the LLM control the generation results?
\end{itemize}

\subsection{Experiment Settings}
\noindent{\bf Datasets.} We experiment on two real-world trajectory datasets, i.e., Chengdu and Porto, which are widely used by previous trajectory generation studies~\cite{difftraj,controltraj,cardiff}. The road networks of the two cities are downloaded using OSMNX~\cite{osmnx} from OpenStreetMap. For both datasets, we use a 90\%/10\% split for training and testing. More details about
the datasets are provided in Appendix~\ref{app:dataset}.

\stitle{Baselines.} We compare HTP with 6 SOTA GPS trajectory generation methods: \textit{TrajGAN}~\cite{trajgan}, \textit{TrajVAE}~\cite{trajvae}, \textit{DiffWave}~\cite{diffwave}, \textit{DiffTraj}~\cite{difftraj}, \textit{ControlTraj}~\cite{controltraj}, and \textit{Cardiff}~\cite{cardiff}. More details about
the baselines are provided in Appendix~\ref{app:baselines}. 

\stitle{Evaluation Metrics.}
Following previous studies~\cite{difftraj,controltraj,cardiff}, we evaluate performance at three levels, i.e., point-level, grid-level, and road-level, using eight metrics to assess the quality of generated trajectories in terms of geometric consistency and movement pattern. The details of evaluation metrics are shown in Appendix~\ref{app:metrics}.

\stitle{Implementation.} HTP is implemented in PyTorch 2.7 and trained on four NVIDIA A100 GPUs. In the first stage, RQ-VAE is trained for 100 epochs using the AdamW~\cite{adamw} optimizer with a learning rate $1\times10^{-4}$ and a cosine scheduler. The per-GPU batch size is 256. The detailed configuration of RQ-VAE is in the Appendix~\ref {app:implement}. In the second stage, we select Qwen3-1.7B~\cite{yang2025qwen3}. Training is conducted with a per-GPU batch size of 16 and 2 epochs. We use the AdamW optimizer with a learning rate $5\times10^{-4}$ and a linear scheduler. The LoRA rank is set to 16. For evaluation, we generate the same number of trajectories as in the test dataset. Following ControlTraj~\cite{controltraj}, we assume that road trajectories are available in the test dataset for evaluation. In practice, road trajectories can be generated using traditional route-planning methods~\cite{route_planning} or existing learning-based road trajectory generation methods~\cite{seed,gtg}.

\subsection{Main Results (RQ1)}
We evaluate the model's performance by three aspects: metric performance, visualization, and length distribution.

\begin{figure*}[!t]
    \centering
    \includegraphics[width=0.9\linewidth]{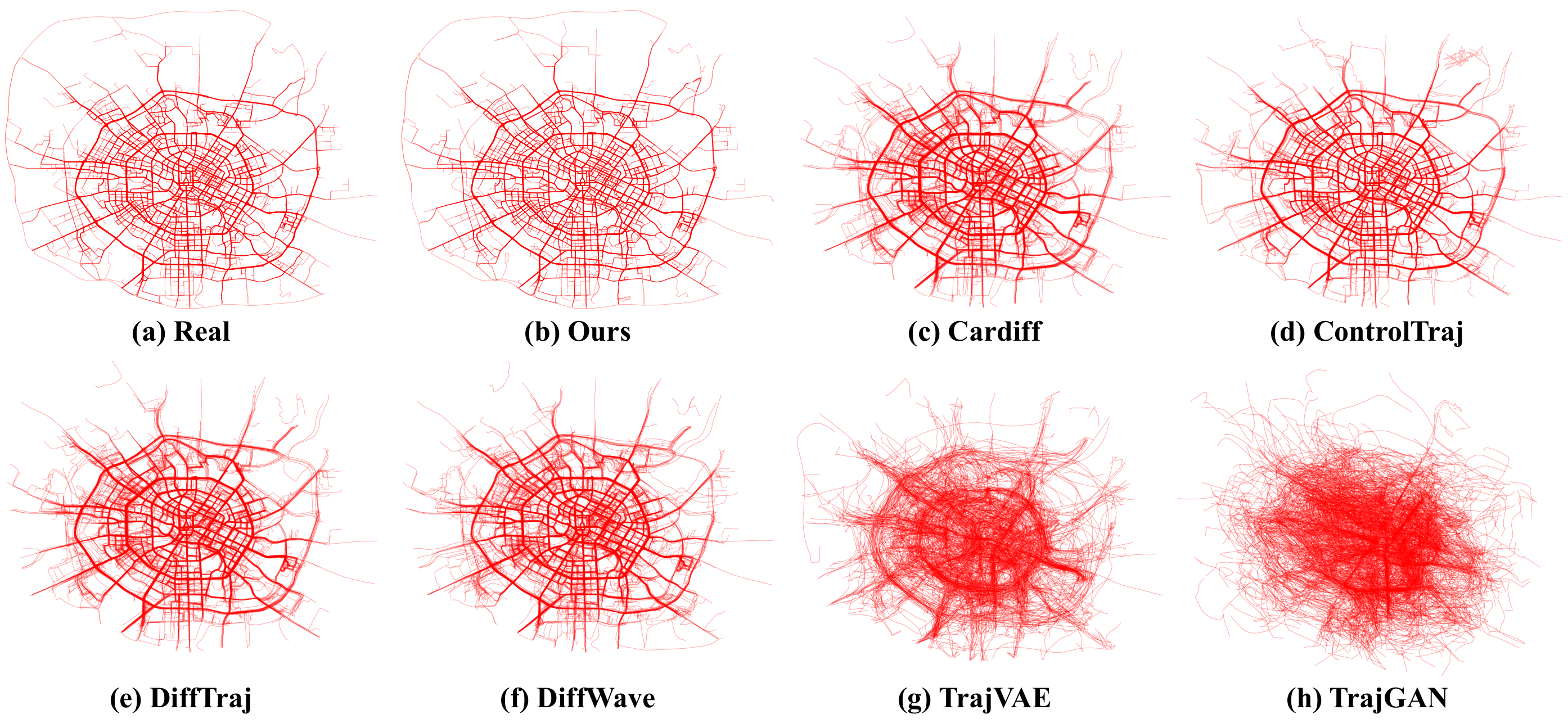}
    \caption{Visualization comparisons on Chengdu dataset.}
    \label{fig:chengdu_traj}
\end{figure*}

\begin{figure*}[!t]
    \centering
    \includegraphics[width=0.9\linewidth]{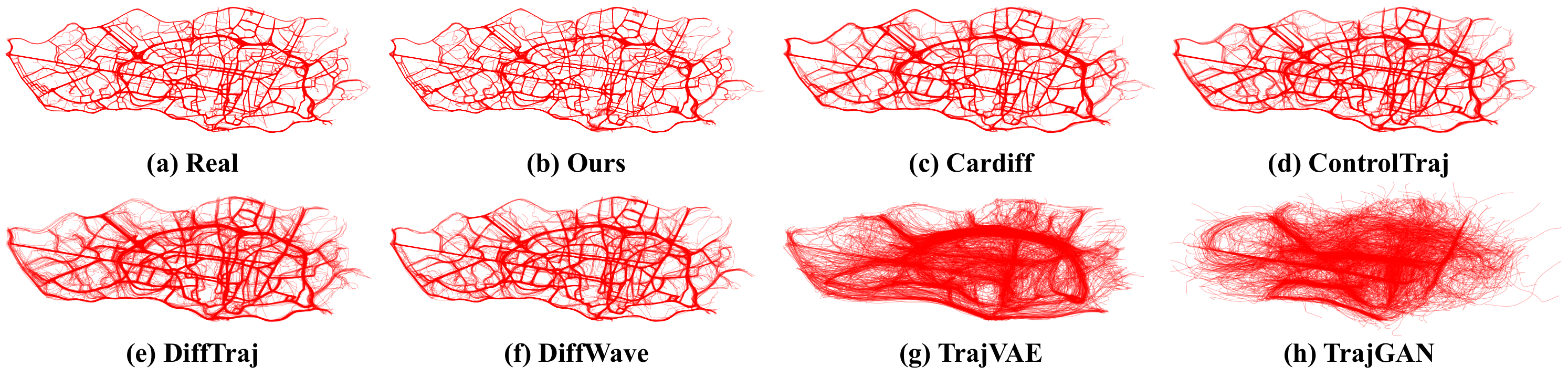}
    \caption{Visualization comparisons on Porto dataset.}
    \label{fig:porto_traj}
\end{figure*}

\begin{figure}[!t]
    \centering
    \includegraphics[width=\linewidth]{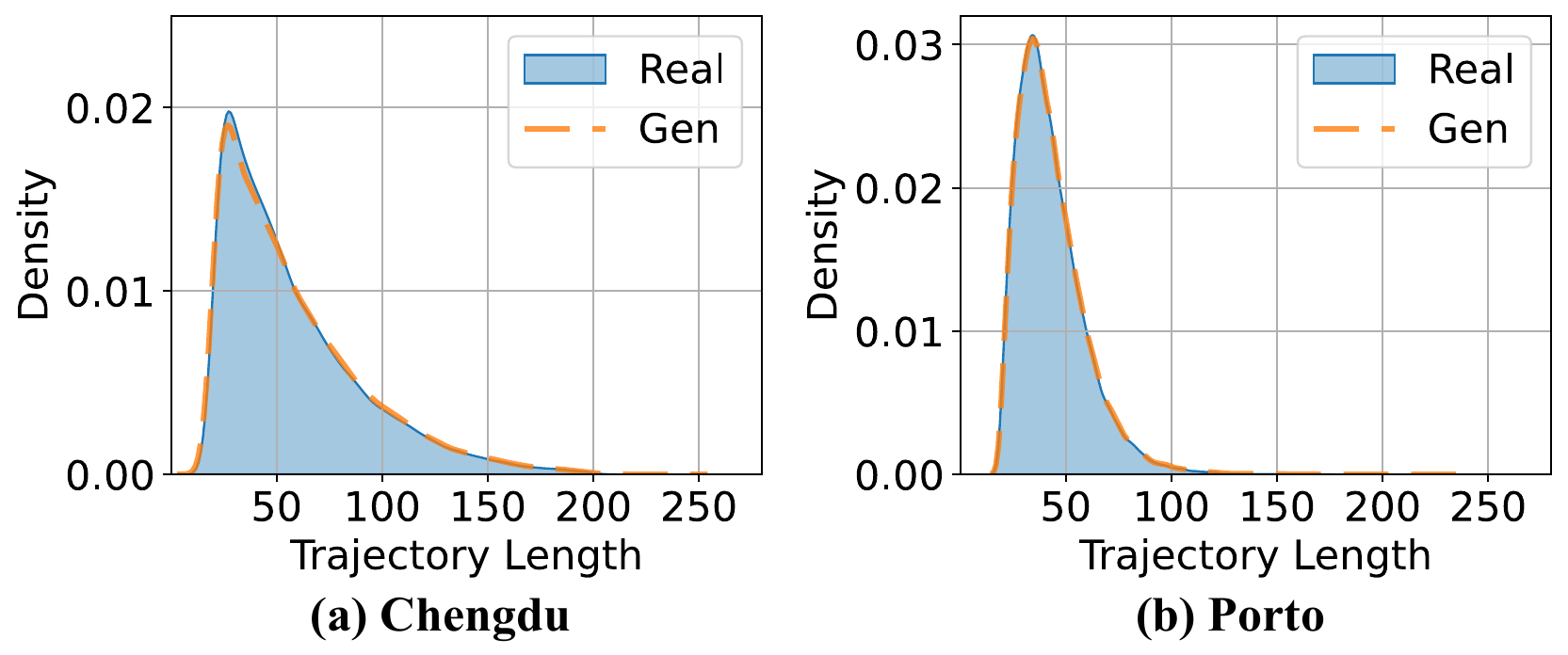}
    \caption{Comparisons of trajectory length density distributions between the real dataset and the generated dataset.}
    \label{fig:length_kde}
\end{figure}

\stitle{Metric Performance.} Table~\ref{tab:main_results} shows the overall performance. We make the following observations.

\begin{table}[!t]
 \LARGE
	\centering
	\caption{Results of HTP when disabling the key designs.}
	\resizebox{\linewidth}{!}{
		\begin{tabular}{l|ccc|ccc}
			\toprule
          & \multicolumn{3}{c}{Chengdu} & \multicolumn{3}{c}{Porto}
          \\\cmidrule(lr){2-4}\cmidrule(lr){5-7}
			& T-Dist$\downarrow$   
            & G-Pat$\uparrow$  
            & R-Pat$\uparrow$  
			& T-Dist$\downarrow$   
            & G-Pat$\uparrow$  
            & R-Pat$\uparrow$  
			\\
			\midrule
            w/o RQ-VAE
            & 0.00932  & 0.73220  & 0.66667
            & 0.00923  & 0.79070  & 0.89630
			\\ 
			w/o R-Loss
			& 0.00244  & 0.09790  & 0.15054  
            & 0.00148  & 0.47934  & 0.65487
			\\   
            w/o OOV
            & 0.00178  & 0.75229  & 0.69058     
            & 0.00542  & 0.64317  & 0.80374
			\\   
            w/o LLM
            & 0.00051  & 0.73973  & 0.91262
            & 0.00354  & 0.69058  & 0.81690
			\\ 
			\midrule
            \textbf{HTP(ours)}
            & \textbf{0.00027}  & \textbf{0.84360}  & \textbf{0.94118}  
            & \textbf{0.00020}  & \textbf{0.83019}  & \textbf{0.92683}
			\\ 
			\bottomrule
		\end{tabular}
	}
	\label{tab:ablation}
\end{table}

First, HTP achieves the best performance across all metrics on both datasets. Specifically, HTP achieves a maximum improvement of 80.51\%, with an average gain of 29.78\%. These results demonstrate that our method can not only capture fine-grained movement patterns at the GPS point level, but also preserve realistic macro-level geographical distributions on grids and roads. At the point level, HTP better matches the fine-grained spatial distributions, indicating more accurate micro-level geometric consistency. At the grid level, HTP more faithfully preserves urban spatial density and high-frequency activity regions, reflecting realistic regional movement patterns. At the road level, HTP generates trajectories that align more closely with road segment usage, follow frequent road patterns, and exhibit smaller point-to-road (PR) distances, demonstrating stronger road-network awareness and travel realism.

Secondly, among existing baselines, diffusion methods, such as ControlTraj and Cardiff, consistently outperform GAN-based and VAE-based methods across both datasets, indicating that diffusion methods are more effective in modeling trajectory distributions. However, despite not using diffusion, HTP still surpasses these diffusion baselines in every metric. Specifically, HTP transforms various conditions into natural language and generates macro-level travel patterns instead of micro-level GPS points, avoiding cumulative coordinate errors while leveraging the world knowledge of LLMs. This demonstrates that the proposed LLM-driven paradigm is a more powerful direction for realistic trajectory generation.

\stitle{Visualization.} Beyond metrics improvements, the visualization comparisons further validate the superiority of HTP. As shown in Figure~\ref{fig:chengdu_traj} and Figure~\ref{fig:porto_traj}, trajectories generated by HTP exhibit clearer adherence to real road structures, more realistic spatial distribution, and smoother motion patterns, closely resembling real-world mobility behaviors. In contrast, baseline methods tend to produce distorted shapes, unrealistic detours, or scattered point distributions. These visual observations clearly confirm that our model not only outperforms existing methods in metrics but also achieves substantially higher fidelity in generating realistic trajectories.

\stitle{Length distribution.} To further validate the ability of HTP to generate realistic variable-length trajectories, we visualize the density distribution of trajectory lengths from generated and real trajectories. As shown in Figure~\ref{fig:length_kde}, the two distributions closely match, demonstrating that HTP effectively supports variable-length generation while preserving real-world length statistics. In contrast, diffusion-based baselines cannot generate variable-length trajectories. TrajGAN and TrajVAE are also excluded, since although they can generate variable lengths, their generated trajectory quality is too poor to make the distribution meaningful.

\subsection{Ablation Study (RQ2)}
To validate the contribution of each component in HTP, we design four ablation variants:
\begin{itemize}[leftmargin=*]
    \item \textbf{w/o RQ-VAE}: Removes the RQ-VAE module and directly fine-tunes the LLM to generate raw GPS trajectories.
    \item \textbf{w/o R-Loss}: Removes the proposed relative reconstruction loss and replaces it with the MSE loss on GPS points.
    \item \textbf{w/o OOV}: Removes the out-of-vocabulary (OOV), including travel pattern, road segment, and length tokens, and relies solely on the base LLM vocabulary.
    \item \textbf{w/o LLM}: Removes the LLM and replaces it with a vanilla Transformer. A sentence Transformer T5~\cite{t5} is used to encode textual descriptions into condition embeddings, which are injected at the first token position.
\end{itemize}

Table~\ref{tab:ablation} demonstrates the effectiveness of each key design in HTP. Removing the proposed relative reconstruction loss (w/o R-Loss) leads to a drastic degradation across all metrics, as the model falls back to unstable trajectory reconstruction in free space, causing severe optimization difficulties. Without the specialized OOV tokens (w/o OOV), the LLM struggles to express numerical travel patterns, road segments, and length information, causing significant declines in both grid-based and road-level pattern accuracy. When RQ-VAE is removed (w/o RQ-VAE) and the LLM directly generates raw GPS points, the performance drops even further because cumulative point errors cannot be controlled. Meanwhile, the lack of token-level compression leads to extremely long generation sequences, which greatly increases inference time. Finally, removing the LLM backbone (w/o LLM) consistently degrades results across both datasets, as it lacks the semantic reasoning capability needed to effectively interpret various conditions, weakening the alignment between conditions and generated trajectories.

\noindent
\textbf{Effect of Road Trajectory.} Considering that the input to the LLM includes road trajectories, which need to be generated by other methods, we further investigate whether the model can work without road-based conditions. To this end, we design a variant by removing all road conditions and letting the LLM first generate road trajectories, followed by travel patterns based on the remaining conditions, denoted as \textbf{HTP+}. The results of the Porto dataset are shown in Table~\ref{tab:htp_wo_road}. We observe that even without road-based conditions, HTP+ can still generate high-quality trajectories.

\begin{table}[!t]
\centering
\caption{Results of HTP when jointly generating road trajectories and travel patterns on the Porto dataset.}
\label{tab:htp_wo_road}
\begin{tabular}{l|c|c|c|c}
\hline
\textbf{Method} & \textbf{T-Dist}$\downarrow$ & \textbf{G-Den}$\uparrow$ & \textbf{R-Den}$\uparrow$ & \textbf{PR-Dist}$\downarrow$ \\
\hline
ControlTraj & 0.00066 & 0.97733 & 0.98003 & 0.06777 \\
Cardiff & 0.00279 & 0.98196 & 0.98641 & 0.05993 \\
\textbf{HTP} & 0.00020 & \textbf{0.99519} & \textbf{0.99418} & 0.01799 \\
\textbf{HTP+} & \textbf{0.00019} & 0.99230 & 0.99106 & \textbf{0.01758} \\
\hline
\end{tabular}
\end{table}

\begin{figure}[!t]
    \centering
    \includegraphics[width=\linewidth]{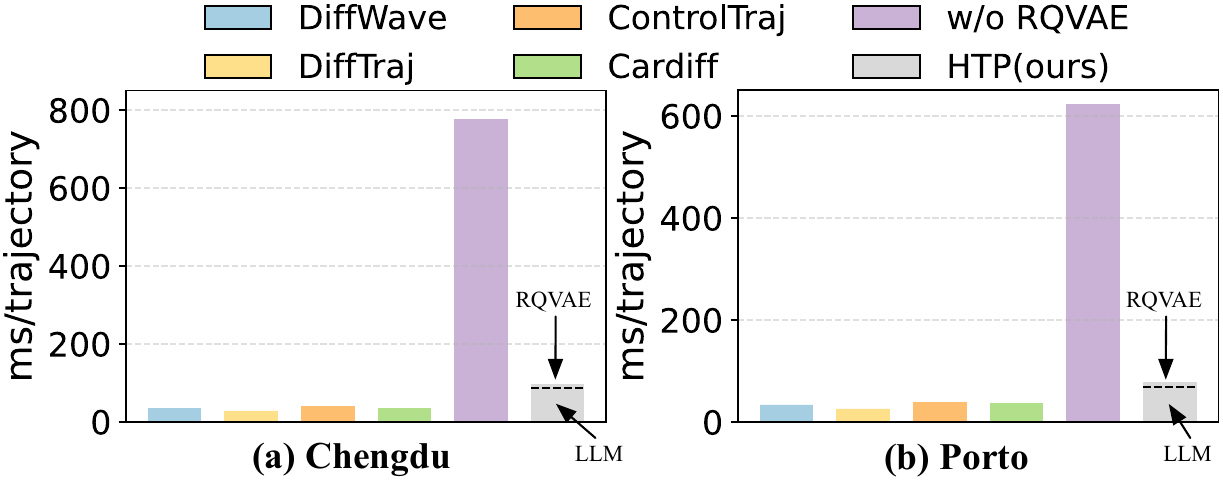}
    \caption{Comparisons of generation speed for one trajectory.}
    \label{fig:efficiency_study}
\end{figure}

\subsection{Efficiency Study (RQ3)}\label{app:inference_time}
Figure~\ref{fig:efficiency_study} compares one trajectory generation speed (in milliseconds) between HTP and diffusion-based baselines on both datasets. Overall, diffusion-based methods are faster than our method. The primary computational cost of HTP comes from the LLM generation stage, while the RQ-VAE decoding introduces only minor overhead. Nevertheless, despite involving an LLM, HTP is not substantially slower than diffusion-based baselines. This is because the trajectory-specific RQ-VAE greatly compresses raw GPS point sequences into short travel pattern tokens, significantly reducing the number of tokens that the LLM needs to generate. In contrast, the variant w/o RQ-VAE, which directly generates GPS points with the LLM, is nearly ten times slower than HTP. In addition, generation on the Chengdu dataset is slower than on Porto, since trajectories in Chengdu have a longer average sequence length, leading to increased computational cost.

\begin{figure*}[!t]
    \centering
    \includegraphics[width=\linewidth]{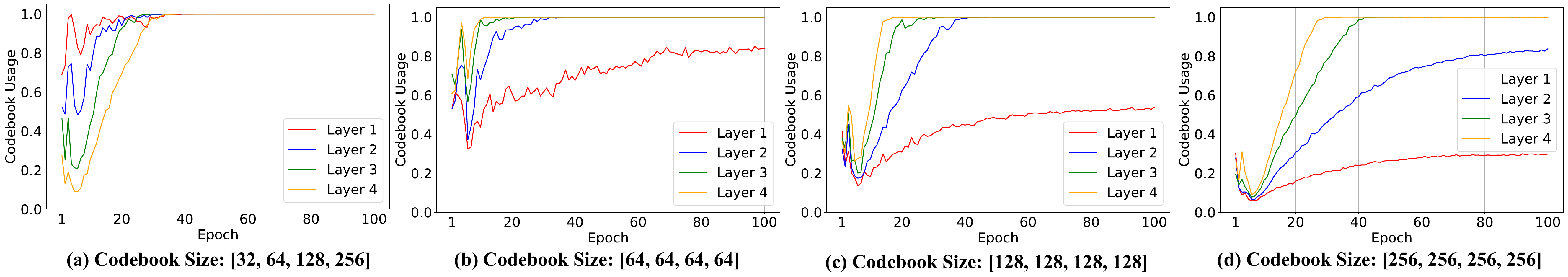}
    \caption{Proportion of token usage in each codebook layer of RQ-VAE during training on the Chengdu dataset.}
    \label{fig:chengdu_usage}
\end{figure*}

\begin{figure}[!t]
    \centering
    \includegraphics[width=\linewidth]{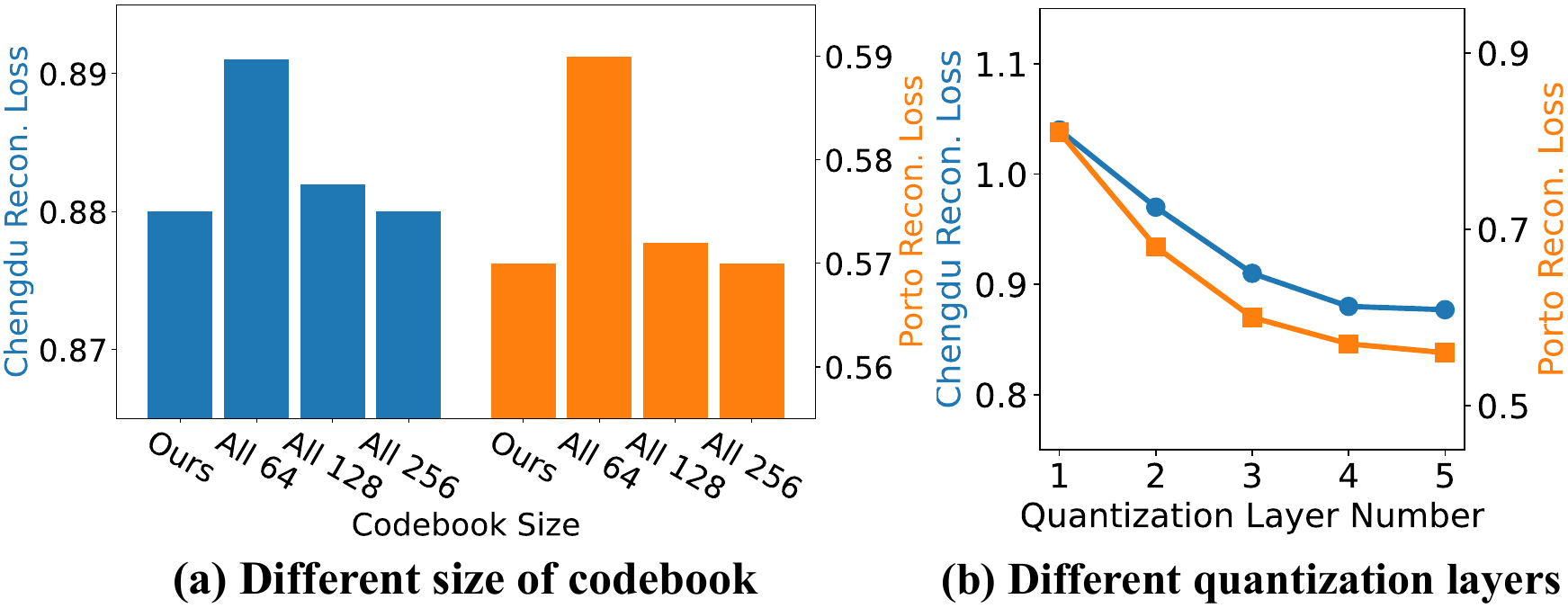}
    \caption{Comparisons of reconstruction loss on codebook size and quantization layer.}
    \label{fig:rq_layer_size}
\end{figure}

\subsection{Hyper-parameter Study (RQ4)}
In HTP, the key hyper-parameters are in the RQ-VAE, including the number of quantization layers and the size of the codebook for each layer. To explore their impact, we further split the training set with a 9:1 ratio into sub-training set and validation set.

\stitle{Codebook Size.} The codebook should be sufficiently large to preserve diversity while its tokens must be well utilized to avoid codebook collapse~\cite{vqvae_collapse}. To find an appropriate setting, we compare validation reconstruction loss. Figure~\ref{fig:rq_layer_size}(a) shows that our setting $\{32, 64, 128, 256\}$ achieves reconstruction performance comparable to the uniformly large setting $\{256, 256, 256, 256\}$, but with lower model complexity. We further analyze token utilization per-layer during training in Figure~\ref{fig:chengdu_usage}. The results reveal: (1) deeper layers converge to a high utilization rate faster, indicating that they capture more fine-grained features; (2) as the codebook size increases, utilization saturates earlier in shallower layers. These suggest that shallow layers mainly encode coarse travel patterns and require smaller codebooks, whereas deeper layers model fine-grained patterns and benefit from larger sizes.

\stitle{Quantization Layers.} We also vary the number of quantization layers from 1 to 5 and report the validation reconstruction loss in Figure~\ref{fig:rq_layer_size}(b). The results show that increasing the number of layers consistently improves reconstruction quality, indicating that multi-resolution quantization effectively captures hierarchical travel patterns. However, the gain saturates after four layers. In particular, on the Chengdu dataset, the reconstruction loss no longer decreases beyond four layers, while a slight further improvement is observed at five layers on the Porto dataset. Nevertheless, considering the marginal gain and the increased computational cost, using four quantization layers offers a good balance between reconstruction accuracy and model efficiency for both datasets.

\begin{figure}
    \centering
    \includegraphics[width=\linewidth]{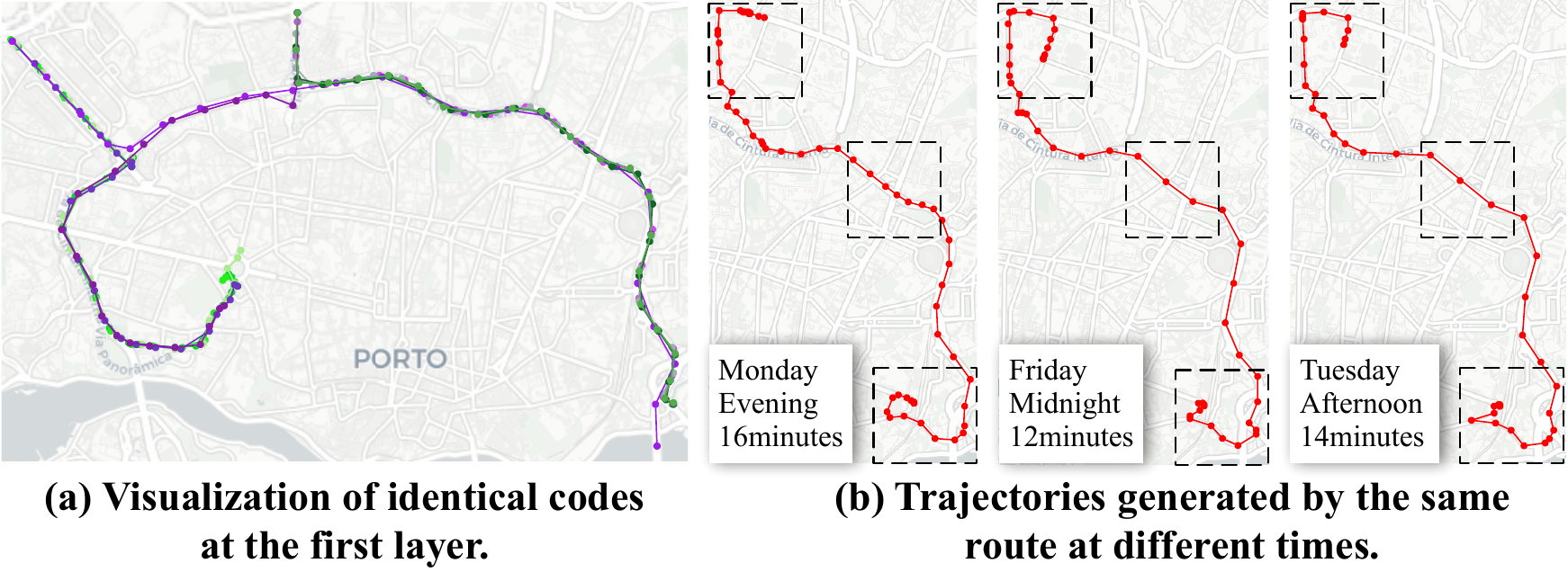}
    \caption{Visualization of case study.}
    \label{fig:case_study}
\end{figure}

\subsection{Case Study (RQ5)} 
\stitle{Codebook.} To further investigate what the codebook has learned, we visualize the trajectories sharing the same codes at the first layer in Figure~\ref{fig:case_study}(a) on the Porto dataset. We observe that these trajectories tend to share similar routes or geometric patterns, such as long straight segments and multiple turns. The point density is further controlled by the deeper layers.

\stitle{Conditional Generation.} Figure~\ref{fig:case_study}(b) presents comparisons of three trajectories generated by HTP for the same route under different time conditions. Although all three samples follow the same road sequence, their spatial point densities vary significantly, reflecting changes in traffic conditions and movement behaviors at different times of day. These variations demonstrate that HTP can modify textual condition descriptions and leverage LLMs to generate different travel patterns, and further produce GPS trajectories with diverse point-density variations by RQ-VAE.

\section{Conclusion}
We propose HTP, a novel framework that hierarchically generates travel patterns first and then generates GPS points. To achieve this, we first design a trajectory-specific RQ-VAE to convert independent micro-level GPS trajectories into shared macro-level travel pattern token sequences. Then we leverage an LLM that is enriched with domain-specific vocabularies and perform supervised fine-tuning to generate travel pattern tokens under various conditions. Extensive experiments demonstrate that HTP can efficiently generate high-quality and realistic variable-length trajectories.

\section*{Acknowledgment}
This paper was supported by the National Key R\&D Program of China 2025YFF0730600, NSFC U25B2049, NSFC U22B2037 and NSFC U25A20443.

\bibliographystyle{ACM-Reference-Format}
\bibliography{ref}

\appendix
\section{Appendix}\label{sec:appendix}

\subsection{Details of Conditions}\label{app:qa_format}
\begin{figure}[t]
    \centering
    \includegraphics[width=0.95\linewidth]{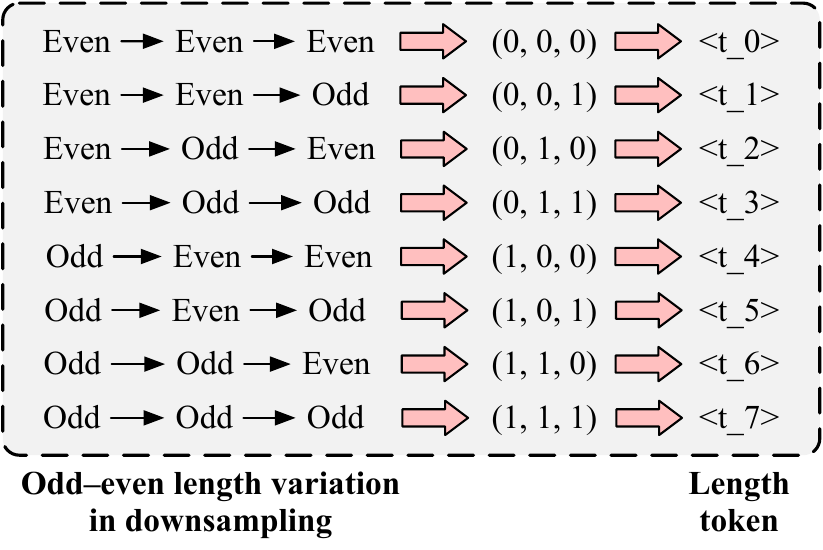}
    \caption{The length token transformation of odd-even variation in downsampling.}
    \label{fig:length_token}
\end{figure}

The input information of the LLM includes road segment tokens describing the road segment IDs, main road segment names, start time, travel time, travel distance, and the GPS sampling interval. The output includes a sampling type (i.e., length token) and a travel pattern sequence. The transformation of odd–even length variation in downsampling into length tokens is shown in Figure~\ref{fig:length_token}.

In practice, during trajectory generation, the road trajectory can be obtained from traditional route-planning algorithms~\cite{route_planning} or existing learning-based road trajectory generation methods~\cite{seed,gtg}. The start time can be freely specified. Travel time can be estimated using average travel durations of road segments, predicted by travel time estimation methods~\cite{travel_time_estimation,travel_time_estimation1}, or manually assigned. The travel distance can be computed from the planned road sequence using road segment length information from the road network.

\subsection{Dataset}\label{app:dataset}
The detailed statistics of the Chengdu\footnote{https://www.pkbigdata.com/common/zhzgbCmptDetails.html} and Porto\footnote{https://www.kaggle.com/c/pkdd-15-predict-taxi-service-trajectory-i} datasets are summarized in Table~\ref{tab:dataset}. The geographic area of Chengdu is approximately ten times larger than that of Porto. During preprocessing, we remove trajectories that fall outside the city boundary or contain fewer than 20 points or more than 200 points. The corresponding road trajectories are obtained via fast map-matching~\cite{fmm}.

\subsection{Baselines}\label{app:baselines}
\begin{itemize}[leftmargin=*]
    \item \textbf{TrajGAN}~\cite{trajgan}: TrajGAN is an encoder–decoder architecture to transform trajectories into dense representations and reconstruct them. A generator learns to produce synthetic trajectory representations from random noise, while a discriminator distinguishes real from generated samples.
    \item \textbf{TrajVAE}~\cite{trajvae}: TrajVAE utilizes an LSTM combined with a variational autoencoder (VAE) to learn trajectory representations and generate new GPS trajectories.
    \item \textbf{DiffWave}~\cite{diffwave}: DiffWave is built upon the Wavenet architecture, and is designed for synthesizing sequences like timeseries, and speech voice. Here we tailor it to generate GPS trajectories.
    \item \textbf{DiffTraj}~\cite{difftraj}: DiffTraj is a diffusion-based trajectory synthesizing framework, which utilizes UNet as the denoising network to generate GPS trajectories.
    \item \textbf{ControlTraj}~\cite{controltraj}: ControlTraj incorporates road trajectories as conditional supervisory signals with the road network topology, which provides road information during the generation of GPS trajectories.
    \item \textbf{Cardiff}~\cite{cardiff}: Cardiff decomposes the generation process into discrete road segment-level and continuous fine-grained GPS-level by the hierarchical diffusion framework.
\end{itemize}

\begin{table}[!t]
\LARGE
	\centering
	\caption{Statistics of the experiment datasets.}
	\label{tab:dataset}
	\resizebox{\linewidth}{!}{
		\begin{tabular}{l|c|c|c|c|c|c}\toprule
			\textbf{Dataset} 
            & \textbf{\#Traj.}
            & \begin{tabular}{c}\textbf{Average}\\\textbf{length}\end{tabular}
            & \begin{tabular}{c}\textbf{Average}\\\textbf{distance}\end{tabular}
            & \begin{tabular}{c}\textbf{Sampling}\\\textbf{interval}\end{tabular}
            & \begin{tabular}{c}\textbf{\#Road}\\\textbf{segment}\end{tabular}
            & \begin{tabular}{c}\textbf{Area}\\\textbf{size}\end{tabular} \\
			\midrule
                Chengdu & 602,880 & 58 points & 4.75km & 10 seconds & 39,503 & 708.6km$^2$  \\
                Porto   & 597,771 & 43 points & 4.83km & 15 seconds & 10,537 & 78.3km$^2$  \\
			\bottomrule
		\end{tabular}
	}
\end{table}

\subsection{Evaluation Metrics}\label{app:metrics}
Following prior work~\cite{difftraj,controltraj,cardiff}, we adopt three categories of evaluation metrics: Jensen–Shannon Divergence (JSD)~\cite{jsd}, cosine similarity, and F1-score. A lower JSD indicates closer alignment between the generated and real statistical distributions. Cosine similarity measures the alignment between two vectors, with higher values indicating greater similarity. The F1-score evaluates the agreement between predicted and ground-truth sets, with higher values indicating better consistency. Using these metrics, we evaluate the similarity between generated and real trajectories at three levels: point-level, grid-level, and road-level.

\begin{figure*}[!t]
    \centering
    \includegraphics[width=\linewidth]{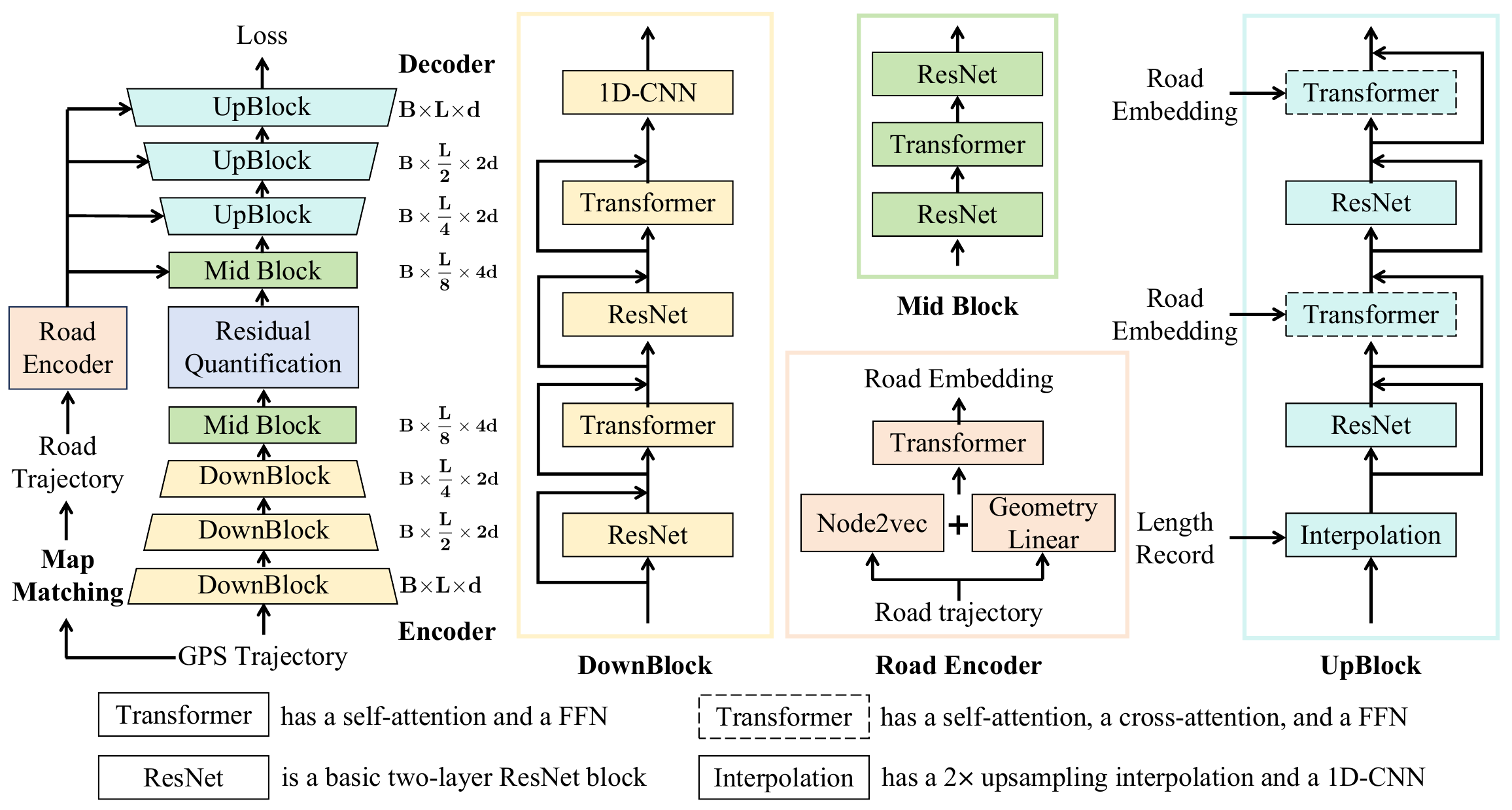}
    \caption{The details of the encoder and decoder of trajectory-specific RQ-VAE.}
    \label{fig:encoder_decoder}
\end{figure*}

\stitle{1. Point-level.}
Point-level metrics evaluate the statistical properties of trajectories at the raw GPS point, focusing on micro-level and fine-grained geometric consistency between generated and real trajectories.

\begin{itemize}[leftmargin=*]
\item \textbf{Travel Distance (T-Dist):}
Measures the JSD on distributions in total travel distances between real trajectories and generated trajectories.
The travel distance of each trajectory is computed as the sum of geodesic distances from the first point to the last point.

\item \textbf{Step Distance (S-Dist):}
Measures the JSD on distributions in step-wise distances between real trajectories and generated trajectories. The step distance is computed as the geodesic distance between two consecutive GPS points.

\item \textbf{Radius:}
Measures the JSD on the spatial range distribution between real trajectories and generated trajectories. Radius is computed by the root mean square distance of all points of a trajectory from their centroid.
\end{itemize}

\stitle{2. Grid-level.}
Grid-level metrics evaluate the spatial aggregation and regional movement patterns of trajectories, capturing how generated trajectories distribute over urban space and whether they preserve high-frequency activity regions. In this paper, each city is partitioned into grids of size $100\text{m} \times 100\text{m}$ to transform GPS trajectories into grid trajectories. 

\begin{itemize}[leftmargin=*]
\item \textbf{Density (G-Den):}
Measures the cosine similarity between the density distribution of the grid aligned by the GPS points of the real trajectory and the generated trajectory.

\item \textbf{Pattern (G-Pat):}
Measures spatial pattern similarity using the F1-score over the top-$k$ most frequently visited grid cells. The top-$k$ grids from real trajectories are treated as ground truth. Here, we set $k=100$.
\end{itemize}

\stitle{3. Road-level.}
Road-level metrics evaluate trajectory realism from a road network perspective, measuring whether generated trajectories follow realistic road usage patterns and align with the underlying road topology. Aligned road segments are obtained via map-matching~\cite{fmm,st_matching}.

\begin{itemize}[leftmargin=*]
\item \textbf{Density (R-Den):}
Measures the cosine similarity between the density distribution of the road segment aligned by the GPS points of the real trajectory and the generated trajectory.

\item \textbf{Pattern (R-Pat):}
Measures road-level pattern similarity using the F1-score over the top-$k$ most frequently visited road segments. The top-$k$ road segments from real trajectories are treated as ground truth. Here, we set $k=100$.

\item \textbf{Point-to-Road Distance (PR-Dist):}
Measures the JSD between the distributions of point-to-road distances, reflecting how well generated trajectories align with the road network.
\end{itemize}

\subsection{Implementation}\label{app:implement}
In the RQ-VAE, we set the latent dimension to $d=64$, the encoder output dimension to $d_e=256$, the quantization dimension to $d_q=64$, and the commitment cost coefficient to $\beta=0.25$. The encoder consists of three downsampling blocks and one mid block, while the decoder includes three corresponding upsampling blocks and one mid block. The dimensionality evolves through the encoder as $64 \rightarrow 128 \rightarrow 128 \rightarrow 256$, with the decoder applying the reverse transformation. Each attention head in both the encoder and decoder has a dimensionality of 64. The Node2vec embedding dimension is 128, and the Transformer for road sequence encoding comprises 4 layers. The quantization module contains 4 layers, with codebook sizes of $[32, 64, 128, 256]$. For the LLM, training on each dataset takes about 15 hours. During trajectory generation, we use the Shapely library\footnote{https://shapely.readthedocs.io/en/stable} to reconstruct GPS trajectories from the predicted relative values.

\end{document}